\documentclass[lettersize,journal]{IEEEtran}
\usepackage{amsmath,amsfonts}
\usepackage{array}
\usepackage{textcomp}
\usepackage{url}
\usepackage{verbatim}
\usepackage{subfigure}
\usepackage{graphicx}
\usepackage{enumitem}
\usepackage{booktabs}
\usepackage[ruled,linesnumbered]{algorithm2e}
\usepackage{xcolor}

\begin{document}
\title{Two-Layer Evolutionary Scheduling for Machine Learning Jobs in Multi-Cluster Environments}
\author{Jingyuan~Xu, 
        Changwu~Huang,~\IEEEmembership{Member,~IEEE,} 
        Lemeng~Pan, 
        Ke~Tang~\IEEEmembership{Fellow,~IEEE}
        and Xin~Yao~\IEEEmembership{Fellow,~IEEE}
\thanks{Jingyuan~Xu and Ke~Tang is with the Guangdong Provincial Key Laboratory of Brain-inspired Intelligent Computation, Department of Computer Science and Engineering of Southern University of Science and Technology, Shenzhen, China.}
\thanks{Changwu~Huang is with the School of AI and Liberal Arts, Beijing Normal-Hong Kong Baptist University, Zhuhai, China.}
\thanks{Lemeng~Pan is with the Huawei Technologies Company, Ltd., Shenzhen, China.}
\thanks{Xin~Yao is with the School of Data Science, Lingnan University, Hong Kong SAR, China.}
\thanks{Corresponding authors: Changwu Huang (changwuhuang@bnbu.edu.cn) and Xin Yao (xinyao@ln.edu.hk).}

}

\markboth{Journal of \LaTeX\ Class Files,~Vol.~14, No.~8, August~2021}%
{Shell \MakeLowercase{\textit{et al.}}: A Sample Article Using IEEEtran.cls for IEEE Journals}


\maketitle

\begin{abstract}
The rapid growth of large-scale machine learning (ML) workloads has led to unprecedented demands on GPU cluster scheduling systems. However, existing heuristic schedulers struggle to handle the complexity and dynamics of multi-cluster GPU environments, often  leading to load imbalance, long queuing delays, and poor alignment {to business} objectives.
This paper formulates the ML scheduling problem in multi-cluster GPU environments as online multi-objective optimization that minimizes the average and maximum weighted queuing times while maximizing GPU time utilization. To {solve this optimization} problem, we propose EVO-GP, a two-layer evolutionary online scheduling method that decouples global job-to-cluster allocation from intra-cluster GPU resource allocation. At the {job-to-cluster} allocation layer,  an evolutionary algorithm assigns arriving jobs to feasible clusters to promote load balance across clusters. At the resource allocation layer, a priority function automatically learned offline via genetic programming is used to rank jobs and guide GPU allocation within each cluster.
Extensive experiments on both real-world {data} and simulated workloads demonstrate that the proposed EVO-GP method consistently outperforms state-of-the-art baselines, while maintaining {superior} adaptability and scalability.
\end{abstract}


\section{Introduction}
\IEEEPARstart{I}{n} recent years, machine learning (ML) techniques, particularly deep learning, have achieved remarkable success across various domains, including speech recognition, robotic control, and medical diagnostics~\cite{survey-dl,survey-dlwsigd}. To meet the growing computational demands associated with ML model training, many leading technology companies have deployed large-scale GPU clusters~\cite{trace-capodlwilsgd,trace-MLaaS,trace-philly}. However, the inherent complexity and distinctive characteristics of ML jobs often lead to low resource utilization and high operational costs under conventional scheduling strategies~\cite{method-mesos,method-yarn}. Consequently, designing efficient scheduling strategies for GPU clusters has become a critical research challenge. 

Existing GPU cluster schedulers can be broadly classified into three categories. The first category is heuristic schedulers~\cite{method-spark,method-liquid,method-tiresias,method-lyra,method-astraea}, which rely on predefined rules (such as First-In First-Out (FIFO)~\cite{method-spark,method-liquid}, Shortest Job First (SJF)~\cite{method-tiresias}) to schedule jobs. While simple, these manually designed heuristics lack flexibility and generalization, making them difficult to adapt to dynamic workloads or changing business objectives. The second category consists of Integer Linear Programming (ILP)-based schedulers~\cite{method-sia,method-osafhdmlj,method-gadget}. However, the strict requirement for linear objectives and constraints restricts their modeling flexibility and scalability in complex real-world environments. The third category encompasses reinforcement learning (RL)-based schedulers~\cite{method-DL2,method-dlbjpidmlc,method-sched2,method-jsflsmlc}, which learn scheduling policies from historical data or online interactions. {While RL can capture dynamic patterns, the resulting policies lack sufficient interpretability and controllability, hindering the incorporation of domain expertise and making system behaviors difficult to explain~\cite{book-rl}}.

From the perspective of scheduling scenarios, most existing GPU cluster schedulers focus on job scheduling within a single cluster. In real-world operations, however, large-scale GPU clusters often suffer from severe inter-job interference caused by resource contention~\cite{method-lsmlcsvmagrl}, which can make shared clusters less efficient than dedicated private clusters~\cite{method-hived}. Consequently, many systems adopt a cluster partitioning strategy, dividing a single large cluster into several logically isolated virtual clusters (VCs) to reduce management complexity and improve resource utilization~\cite{trace-capodlwilsgd,trace-philly}. While this approach mitigates interference, it can lead to load imbalance across clusters: when some VCs experience serious queuing delays due to job backlogs, other VCs may remain underutilized, causing resource wastage. A common remedy is to periodically adjust the size of each VC, scaling down those under low {loads} and scaling up those under high {demands}~\cite{trace-capodlwilsgd}. However, such reconfigurations are typically performed only every few days~\cite{trace-capodlwilsgd}, which is insufficient to respond to real-time workload fluctuations and limits the benefit of multi-cluster coordination.

These observations reveal two key {research} gaps. From the method perspective, existing schedulers are either rigid (heuristics and ILP approaches) or opaque (RL-based methods), making them difficult to jointly optimize multiple objectives while retaining interpretability and ability to incorporate domain knowledge. From the scenario perspective, most work either {focused} on single-cluster environments or handle multi-cluster settings in an ad hoc and coarse-grained manner, lacking principled mechanisms for online, fine-grained coordination across clusters.

To address these challenges, this {paper} formulates ML job scheduling in multi-cluster GPU environments as a multi-objective optimization problem that minimizes the average and maximum weighted queuing times while maximizing {GPU time utilization}. {Then}, we propose EVO-GP, a two-layer evolutionary online scheduling framework for the formulated ML training job scheduling in multi-cluster GPU environments. {This new framework} adopts a collaborative optimization mechanism that decouples global job allocation from local resource allocation. At the job allocation layer, an evolutionary algorithm dynamically assigns incoming jobs to suitable clusters to ensure global load balance across clusters. At the resource allocation layer, a priority function is automatically learned by genetic programming (GP) and used to determine job priorities and {guide} GPU allocation within each cluster. {Such a framework} combines the global search capability of evolutionary algorithms with the interpretability of GP-learned rules.

The main contributions of this paper {include}:
\begin{itemize}
\item \textbf{Problem formulation for multi-cluster ML job scheduling.} 
{Unlike existing scheduling models that are confined to a single cluster}, we explicitly formulate the scheduling problem for rigid ML jobs in heterogeneous multi-cluster environments. The objectives minimize average and maximum weighted queuing times while maximizing resource utilization, aligning with both efficiency and practical requirements.

\item \textbf{Two-layer evolutionary online scheduling method.} We propose EVO-GP, a {novel} two-layer evolutionary online ML job scheduling method that {performs} job allocation and resource allocation. The upper layer handles global load balancing via evolutionary search, while the lower layer manages local resource {allocation} using a learned priority function.

\item \textbf{Priority function automatically learned by GP.} We employ a GP-based approach to automatically discover high-quality, human-interpretable priority functions. This offline learning process reduces reliance on {human}-designed heuristics and improves adaptability to varying optimization preferences.

\item \textbf{Comprehensive evaluation and generalization.} Extensive experiments on both real-world traces and simulated workloads, in both multi-cluster and single-cluster settings, demonstrate that the proposed EVO-GP consistently outperforms state-of-the-art baselines {according to commonly used} key metrics. It balances multiple objectives and generalizes well across different cluster {environments}.
\end{itemize}

The remainder of this paper is organized as below. Section~\ref{sec:background} formalizes the multi-cluster ML job scheduling problem and briefly reviews the related work. Section~\ref{sec:evo-gp} introduces our proposed EVO-GP framework, detailing the evolutionary job allocation algorithm and the GP-based priority function learning method. Section~\ref{sec:exp} presents the experimental setup and analyzes the results. Finally, Section~\ref{sec:conclusion} concludes the paper.

\section{Background}
\label{sec:background}
In this section, we first formalize the online ML job scheduling problem in a multi-cluster GPU environment. We then review existing scheduling methods for ML workloads in both single-cluster and multi-cluster settings, highlighting their design principles and limitations. This background lays the foundation for our proposed scheduling framework and clarifies how it differs from and advances prior work.

\subsection{Problem Formulation}
In this subsection, we formalize the online ML job scheduling problem in a multi-cluster GPU environment. We first present the system model and job attributes, then describe the online scheduling dynamics, and finally define the performance objectives and constraints that together constitute our problem formulation.

\subsubsection{System Model}
\label{model}
We consider a GPU system that consists of $P$ clusters, denoted as $\mathcal{P} = \{1, 2, \dots, P\} $. Each cluster $p \in \mathcal{P}$ contains $N_p$ homogeneous nodes, {denoted as $\mathcal{N}_p=\{1, 2, \dots, N_p\}$}, each equipped with $ G_p $ GPUs. Thus, the total number of GPUs in cluster $p$ is $N_p \times G_p$. This multi-cluster GPU environment serves as the foundational resource substrate for ML job scheduling~\cite{trace-capodlwilsgd, method-astraea}.

The set of jobs is denoted as $\mathcal{J} = \{1, 2, \dots, J\}$. Each job $ j \in \mathcal{J} $ is characterized by its arrival time $a_j$, {a} user-specified cluster $s_j$ to which it is initially submitted, and an (estimated) training time or duration $d_j$, {the} required number of workers $r_j$ and a total GPU demand $g_j$, and {a} user-defined priority $u_j$, where a larger $u_j$ indicates a more urgent job. 
{Here, a worker represents an {execution unit (e.g., a process in a virtual machine or container) on a compute node that executes a portion of the distributed training workload}~\cite{method-osafhdmlj}}. Following prior works~\cite{method-tiresias,method-jsflsmlc}, we assume that the training time $d_j$ of each job $j$ is known. In practical applications, $d_j$ can be replaced {by an} estimated duration of the job~\cite{method-optimus}. Each worker of job $j$ {is} assumed to consume $\frac{g_j}{r_j}$ GPUs. These attributes jointly determine the resource demand and urgency of each job, and are crucial for designing effective scheduling strategies.

When the scheduler performs a scheduling operation, it involves two  key decisions: {(1) Job allocation across clusters.} For each cluster $p$, the set of jobs tentatively assigned to cluster $p$ is denoted by $\mathcal{Y}_p$. To balance load, a job $j$ may be reassigned to a cluster different from its user-specified cluster $s_j$; the actual allocated cluster is denoted by {$c_j$}. {(2) Resource allocation within clusters.} For each cluster $p$, this decision jointly determines: {(i) Priority ordering within a cluster.} For each cluster $p$, the relative priority of jobs in $\mathcal{Y}_p$ is represented by a permutation (vector) $\mathbf{y}_p$. A job that appears earlier in $\mathbf{y}_p$ is assigned a higher ``relative priority". We use ``relative priority" to denote the ordering produced by a scheduling rule (e.g., FIFO, SJF) or a learned priority function, so as to distinguish it from the user-defined priority $u_j$. {(ii) Worker placement on nodes.} Let $x_{j,n} \in \{0,1,\ldots,r_j\}$ denote the number of workers of job $j$ placed on node $n$. If $x_{j,n}=0$, then job $j$ is not executed on node $n$. The placement variables $x_{j,n}$ must satisfy both job-level and node-level capacity constraints.

\subsubsection{Online ML Job Scheduling in {a} Multi-Cluster Environment}
We now describe the online scheduling problem in a dynamic setting. Before making a scheduling decision at time $t$, the scheduler must obtain both static and dynamic information about jobs. In addition to the static attributes introduced in Section \ref{model}, at time $t$, each job $j$ is associated with a remaining training time {$d_j^{(t)}$} and an already accumulated queuing time {$q_j^{(t)}$}. Thus, in the dynamic model, a ML job can be represented as an octuple $j=\langle a_j, s_j, d_j, r_j, g_j, u_j, d_j^{(t)}, q_j^{(t)}\rangle$, where the first six components are static and the last two capture the current execution state. Let $\mathcal{J}^{(t)}$ denote the set of jobs that have arrived by time $t$ but have not yet completed. This set is partitioned into running jobs $\mathcal{J}_e$ and queued jobs $\mathcal{J}_q$. For each running job $j\in \mathcal{J}_e$, let {$w_{j,n}$} denote the number of workers of job $j$ placed on node $n$ before the scheduling operation at time $t$.

Given a scheduling decision at time $t$—including the job-to-cluster assignment $\{\mathcal{Y}_p\}$, the relative priority order $y_p$ within each cluster, and the worker placement $\{x_{j,n}\}$—we can derive the start time of each job. Let $S_j$ denote the start time of job $j \in \mathcal{Y}_p$, where $S_j \geq t$. If the resources are sufficient at time $t$, i.e., $\sum_{j \in \mathcal{Y}_{p,\leq j}} x_{j,n} \cdot \frac{g_j}{r_j} \leq G_p$ holds for all $n \in \mathcal{N}_p$ (where $\mathcal{Y}_{p,\leq j}$ is the set containing job $j$ and all jobs in $\mathcal{Y}_p$ that have a higher relative priority than $j$), then job $j$ can start immediately and thus $S_j = t$. Otherwise, job $j$ must wait until some higher-priority jobs complete and release sufficient resources. In this case, the start time of job $j$ is
\begin{equation}
\begin{aligned}
    S_j= \min\limits_{w\in\mathcal{Y}_{p,< j}}\{F_w|\sum\limits_{k\in\mathcal{Y}_{p,\leq j}}(x_{k,n}\cdot&\frac{g_k}{r_k}\cdot\mathbf{1}_{F_k>F_w})\\ &\leq G_p, \forall n\in\mathcal{N}_p\},
\end{aligned}
\label{start_time}
\end{equation}
where $F_w = S_w + d^{(t)}_w$ denotes the completion time of job $w$, $\mathbf{1}_{F_k>F_w}$ is an indicator function that takes the value $1$ if $F_k > F_w$ and $0$ otherwise, {and $w \in \mathcal{Y}_{p,< j}$ denotes the set of jobs in $\mathcal{Y}_p$ that have a higher relative priority than $j$}.
 The queuing time of job $j$ under the current scheduling scheme is denoted by $q_j^{(t)}=q_j^{(t)} + S_j-t$, which consists of the time job $j$ has already spent waiting before time $t$ plus the additional waiting time from $t$ until its start time $S_j$.

\subsubsection{Objective Functions}
We consider three objectives, and the specific calculation methods for each objective are as follows.

\paragraph{Average Weighted Queuing Time}
Following~\cite{method-osafhdmlj}, we define the average weighted queuing time as
\begin{equation}
C_{q,avg}=\frac{1}{|\mathcal{J}^{(t)}|}\sum_{j\in\mathcal{J}^{(t)}}u_j\cdot q_j^{(t)}
\label{AvgWQT}
\end{equation}
Here, the user priority $u_j$  serves as a weight for the queuing time of job $j$. This metric encourages earlier execution of jobs with {a} higher user priority.

\paragraph{Maximum Weighted Queuing Time}
Classical heuristic algorithms, such as the Shortest Remaining Service First (SRSF) algorithm~\cite{method-tiresias}, can achieve favorable performance in terms of average job completion time. However, they tend to prioritize jobs with shorter remaining times and lower resource requirements, which may cause large jobs to experience excessively long queuing delays, degrading user experience. To alleviate this issue and avoid starvation, we introduce the maximum weighted queuing time:
\begin{equation}
C_{q,max}= \max_{j\in\mathcal{J}^{(t)}} u_j\cdot q_j^{(t)}
\label{MaxWQT}
\end{equation}

\paragraph{GPU Time Utilization} 
To reduce resource fragmentation and improve GPU resource efficiency, we define GPU time utilization as the ratio between the total GPU hours requested by jobs and the total GPU capacity over the planning horizon:
\begin{equation}
C_{u}= \frac{\sum_{j\in \mathcal{J}^{(t)}} g_j\cdot d_j^{(t)} }{\left( \sum_{p\in\mathcal{P}} G_p\cdot N_p \right) \cdot \left( \max\limits_{j\in\mathcal{J}^{(t)}} F_j - t \right)}
\label{utilization}
\end{equation}

\paragraph{{Combined} Objective Function of Online Multi-Objective Scheduling Problem}
We combine the three objectives via weighted-sum and formulate the following minimization problem:
\begin{equation}
    {\rm minimize}\quad \omega_1\cdot C_{q,avg} + \omega_2\cdot C_{q,max} + \omega_3 \cdot (1-C_{u})
\label{objective}
\end{equation}
${\text{\quad \quad subject to}}$
\begin{align}
    &\mathcal{Y}_1\cap\mathcal{Y}_2\cap ... \cap \mathcal{Y}_{P}=\emptyset\label{c6}\\
    &\mathcal{Y}_1\cup\mathcal{Y}_2\cup ... \cup \mathcal{Y}_{P}=\mathcal{J}^{(t)}\label{c7}\\
    &\sum_{n\in\mathcal{N}_p}x_{j,n}=r_j,\forall p\in\mathcal{P}, \forall j\in\mathcal{Y}_{p}\label{c8}\\
    &\sum_{j\in\mathcal{Y}_p} x_{j,n}\times \frac{g_j}{r_j}\leq G_p,\forall p\in\mathcal{P}, \forall j\in\mathcal{Y}_{p}\label{c9}\\
    &\left\lfloor\frac{G_p\times r_j}{g_j}\right\rfloor\times N_p\geq r_j,\forall p\in \mathcal{P},\forall n\in\mathcal{N}_p,\forall j\in\mathcal{Y}_p\label{c10}\\
    &S_j=t\wedge x_{j,n}=w_{j,n}, \forall p\in\mathcal{P},\forall n\in\mathcal{N}_p, \forall j\in\mathcal{J}_n \label{c11}
\end{align}

We employ weighted-sum scalarization to integrate the objectives into a single optimization target. Constraints \eqref{c6} and \eqref{c7} enforce unique job allocation, i.e., each job is assigned to exactly one cluster and cannot simultaneously occupy resources in multiple clusters. Constraint \eqref{c8} ensures that the number of workers placed for each job equals its required worker count. Constraint \eqref{c9} is the node capacity constraint, requiring that the total number of GPUs used by all jobs on any node does not exceed the node’s GPU capacity. Constraint \eqref{c10} is the cluster capacity constraint, ensuring that the number of workers required by a job $j$ does not exceed the maximum number of workers that cluster $p$ can host for that job. Here, $\left\lfloor\frac{G_p \times r_j}{g_j}\right\rfloor$ denotes the maximum number of workers of job $j$ that a single node in cluster $p$ can accommodate. Constraint \eqref{c11} defines the preemption constraint for jobs. Specifically, a non-preemptive job refers to one whose allocated resources remain unchanged under the new scheduling scheme and does not need to re-enter the queue. The set $\mathcal{J}_n$ denotes the set of non-preemptive jobs. In practice, whether a job can be preempted is determined by specific rules, which may vary across different business scenarios. In this paper, we employ a simple rule to define the set of non-preemptive jobs as $\mathcal{J}_n = \{j\in\mathcal{J}_e \mid s_j = c_j\}$, which implies that jobs currently running in the user-specified cluster cannot be preempted. Our work focuses on the scheduling decision process rather than the system-level overhead associated with job preemption. Therefore, we assume that preempted jobs can be resumed without any loss of progress.

These definitions complete our formulation of the online multi-cluster ML job scheduling problem. We next review existing scheduling methods in both single-cluster and multi-cluster environments.

\subsection{Related Work}

In this subsection, we review existing scheduling methods for ML workloads that are most relevant to our problem.

{Most existing studies focus on single-cluster scheduling.  They employ heuristic rules~\cite{method-tiresias,method-liquid,method-lyra,method-astraea}, mathematical programming~\cite{method-osafhdmlj,method-sia,method-gadget}, or reinforcement learning~\cite{method-deeprm,method-sched2,method-jsflsmlc,method-DL2} to resolve resource contention within a single cluster. }Despite their effectiveness, these classical scheduling methods become increasingly inadequate as cluster scale grows. As cluster size expands, contention on shared resources (e.g., network and storage) can significantly degrade training performance~\cite{method-lsmlcsvmagrl}, and a single shared cluster may even underperform dedicated clusters~\cite{method-hived}. Accordingly, several works propose unified schedulers for large-scale clusters~\cite{trace-MLaaS,method-jsflsmlc,method-pop,method-lsmlcsvmagrl}. In industry, large clusters are often partitioned into multiple isolated virtual clusters (VCs) to simplify management and mitigate interference~\cite{trace-capodlwilsgd,trace-philly}. This partitioning introduces a key challenge: load imbalance across VCs, which can cause resource idleness in some VCs and long queuing delays in others~\cite{trace-capodlwilsgd}. Dynamic VC resizing can partially mitigate imbalance~\cite{trace-capodlwilsgd}, but it is typically performed on multi-day cycles and thus cannot respond to short-term demand fluctuations. Lyra~\cite{method-lyra} addresses imbalance via resource loaning between an inference and a training cluster, but its mechanism is tailored to a two-cluster setting and does not directly generalize to many training clusters. 

{In a multi-cloud task scheduling scenario, Panda \emph{et al.}~\cite{method-multicloud} adopts a heuristic Minimum Completion Time (MCT) policy, which assigns each task to the cluster that yields the smallest predicted completion time. Similarly, MAST~\cite{method-MAST} globally schedules large-scale ML training jobs across geo-distributed datacenters by jointly optimizing placement and data transfer under WAN constraints. However, these approaches are limited by their focus on specific infrastructure constraints and lack the generality and automated policy-discovery capabilities.}

In summary, existing scheduling methods either operate within a single cluster or handle multi-cluster environments in a coarse-grained manner, {typically through infrequent mechanisms such as cluster resizing or limited resource borrowing. Such approaches react to load imbalance only at a coarse time scale and are unable to cope with highly dynamic workload variations. In contrast, we address multi-cluster scheduling by directly and online assigning each arriving job to an appropriate cluster, enabling fine-grained and immediate responses to dynamic load changes.}

Moreover, existing methods are methodologically either human-designed with limited adaptability (heuristics and ILP approaches) or black-box (RL-based methods). This paper formulates multi-cluster ML job scheduling as an online multi-objective optimization problem, {jointly considering average and maximum weighted queuing times while maximizing GPU time utilization.} Building on this formulation, we propose a two-layer evolutionary scheduling framework with GP-learned, human-interpretable priority functions, providing adaptive performance while retaining explicit and controllable scheduling policies in multi-cluster GPU environments.

\section{Two-layer Evolutionary Scheduling for ML Jobs in Multi-Cluster Environments}
\label{sec:evo-gp}
This section presents our proposed two-layer evolutionary online scheduling framework for the formulated ML training jobs in multi-cluster GPU environments. The framework, termed EVO-GP, adopts a collaborative optimization mechanism that explicitly decouples global job allocation from local resource allocation. At the job allocation layer, an evolutionary algorithm dynamically assigns incoming jobs to suitable clusters to ensure global load balance across clusters. At the resource allocation layer, a priority function is automatically learned offline by GP and used to determine job priorities and guide GPU allocation within each cluster. Section \ref{sec:evo-gp-fromework} presents the overall framework of the EVO-GP. Section \ref{sec:job_allocation} describes the implementation details of the evolutionary job allocation, and Section \ref{sec:resource_allocation} introduces the implementation details of the resource allocation procedure and offline learned priority function {using} GP.

\subsection{Algorithm Overview}
\label{sec:evo-gp-fromework}
\begin{figure}[hbtp]
    \centering
    \includegraphics[width=0.4\textwidth]{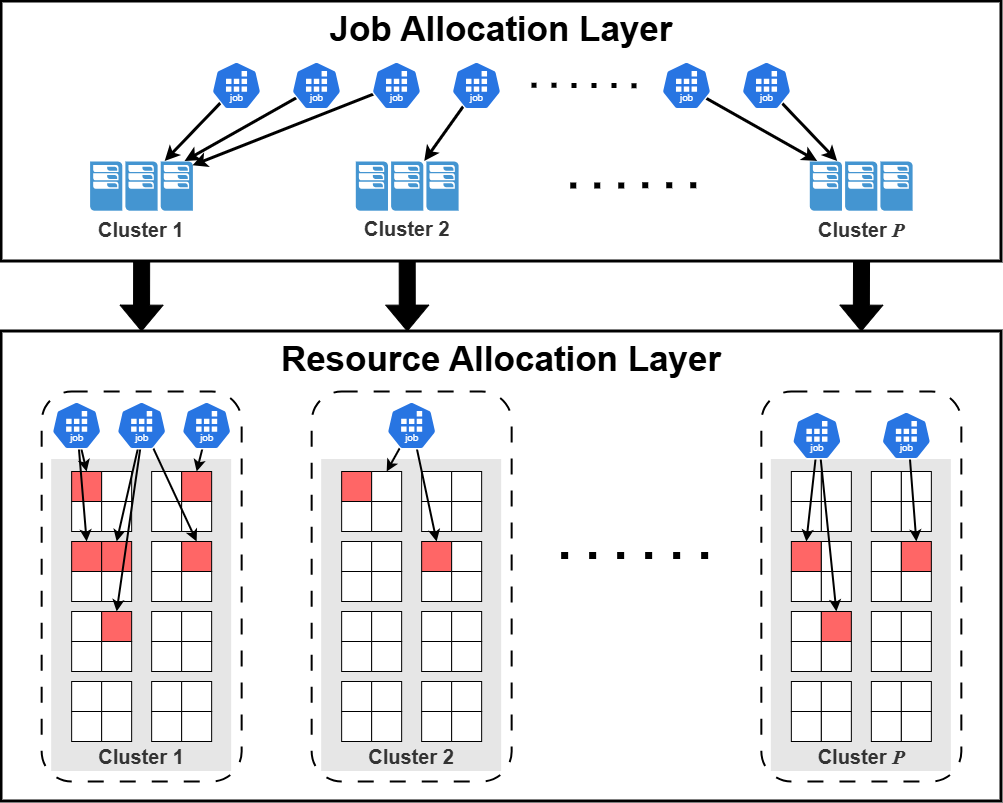} 
    \caption{Overall illustration of EVO-GP.}
    \label{frame}
\end{figure}

EVO-GP is a two-layer evolutionary online scheduling framework that decouples global job allocation from local resource allocation in multi-cluster GPU environments. As shown in Fig.~\ref{frame}, the upper layer (Evolutionary Job Allocation) decides \emph{which cluster each job should run on}, while the lower layer (Resource Allocation with a GP-learned priority function) decides \emph{how to allocate GPUs within each cluster}. This separation addresses the combinatorial complexity of multi-cluster scheduling by (i) searching over job-to-cluster mappings under cluster-level feasibility constraints and (ii) applying a reusable intra-cluster allocation policy. Meanwhile, the priority function used in the resource allocation layer is learned offline by GP, thereby reducing reliance on handcrafted heuristics and enabling adaptation to multi-objective requirements such as delay, fairness, and utilization.

At the job allocation layer, EVO-GP employs an evolutionary algorithm in which each individual represents a complete job-to-cluster assignment. Starting from a feasibility-aware randomized initialization, the algorithm repeatedly applies crossover and mutation to generate new assignments that satisfy cluster feasibility constraints. For each candidate assignment, the lower-layer resource allocation procedure constructs a feasible schedule and evaluates its fitness using the scalarized multi-objective function (Eq.~\eqref{objective}). Through environmental selection, high-quality assignments are retained and guide subsequent search. In this way, the evolutionary job allocation layer intelligently explores the combinatorial space of job–cluster assignments under complex constraints, while remaining agnostic to the internal details of the resource allocation policy.

At the resource allocation layer, EVO-GP performs resource allocation with a GP-learned priority function. Given a fixed job-to-cluster assignment from the upper layer, this module first respects non-preemptive jobs by keeping their current placements and assigning them the highest priority, and then uses the learned priority function $f(\cdot)$ to rank the remaining jobs within each cluster. Jobs are scheduled in the order determined by $f(\cdot)$, and their workers are placed using a Best Fit strategy based on {the} earliest start time and residual GPU capacity, turning a job allocation plan into a complete executable schedule. The priority function $f(\cdot)$ is obtained offline by GP, which searches the space of interpretable expressions over key job features (e.g., remaining duration, queuing time, user priority, GPU demand) to find {a} suitable priority for the scheduling goal. During online scheduling, the evolutionary job allocation layer and the priority-based resource allocation layer collaborate as follows: the upper layer proposes different job-to-cluster mappings, the lower layer realizes these mappings into concrete schedules, and the resulting objective values feed back to the evolutionary search. This collaborative mechanism enables joint optimization of global load balancing and local scheduling quality, while preserving flexibility and interpretability (via a pluggable $f(\cdot)$ in human-readable GP expression).

The implementation details of EVO-GP are shown in Algorithm~\ref{alg:evo-gp}. The inputs include the set of unfinished jobs $\mathcal{J}^{(t)}$, the non-preemptive job set $\mathcal{J}_n$, the cluster set $\mathcal{P}$, the population size $n$, the mutation rate $r$, and the priority function $f(\cdot)$. The algorithm first identifies schedulable jobs $\mathcal{J}_s=\mathcal{J}^{(t)}\setminus\mathcal{J}_n$ (line 1) and encodes them for compact representation (line 2). It then initializes a feasible population (line 3) and iteratively generates offspring via crossover and mutation (lines 5--13). Each offspring individual is evaluated by invoking the resource allocation procedure to construct a complete schedule and compute its fitness, {where $\mathcal{X}_i$ records the worker placements and $\mathcal{Y}_i$ records the intra-cluster job orderings.} (lines 14--17). The best $n$ individuals are selected to form the next generation (line 18). After termination, the best individual is decoded and passed to the resource allocation procedure to output the final schedule (lines 20--22).

\begin{algorithm}[hbtp]
  \SetKwData{Left}{left}\SetKwData{This}{this}\SetKwData{Up}{up}
  \SetKwFunction{Union}{Union}\SetKwFunction{FindCompress}{FindCompress}
  \SetKwInOut{Input}{Input}\SetKwInOut{Output}{Output}

  \Input{Job set $\mathcal{J}^{(t)}$, Non-preemptive job set $\mathcal{J}_n$, Cluster set $\mathcal{P}$, Population size $n$, Mutation rate $r$, Priority function $f(\cdot)$.}
  \Output{Solution $s$.}
  $\mathcal{J}_s\leftarrow \mathcal{J}^{(t)}\setminus \mathcal{J}_n$ \;
  Encode the jobs in $\mathcal{J}_s$\;
  $\mathcal{I}\leftarrow$ Population Initialization\;
   \While{termination conditions are not met}{
        $\mathcal{N}\leftarrow\emptyset$\;
        \tcp{Job Allocation}
        \While{$|\mathcal{N}|<n$}{
        $p_1, p_2 \leftarrow$ Randomly select two individuals from $\mathcal{I}$\;
        $o_1, o_2 \leftarrow$ Perform crossover between $p_1$ and $p_2$\;
        \If{$rand(0,1)< r$}{
            Apply mutation to individuals $o_1$ and $o_2$\;
        }
    
        $\mathcal{N}\leftarrow\mathcal{N}\cup\{o_1,o_2\}$\;
        }
        \tcp{Resource Allocation}
        \For{$i\in\mathcal{N}$}{
            {$(\mathcal{X}_i, \mathcal{Y}_i)\leftarrow$ ResourceAllocation($i$, $\mathcal{J}_s$, $\mathcal{J}_n$, $\mathcal{P}$, $f(\cdot)$)\;}
        }
        Evaluate the fitness of all individuals in $\mathcal{N}$\;
        $\mathcal{I} \leftarrow$ Select $n$ best individuals from $\mathcal{N} \cup \mathcal{I}$\;

   }
    $o\leftarrow$ The optimal individual from $\mathcal{I}$\;
    $(\mathcal{X}, \mathcal{Y})\leftarrow$ ResourceAllocation($o$, $\mathcal{J}_s$, $\mathcal{J}_n$, $\mathcal{P}$, $f(\cdot)$)\;
    $s\leftarrow(\mathcal{X}, \mathcal{Y} )$
  \caption{Pseudo-code of the Proposed Two-layer Evolutionary Scheduling Method (EVO-GP)}
  \label{alg:evo-gp}
\end{algorithm}

\subsection{Details for Evolutionary Job Allocation}
\label{sec:job_allocation}
This subsection details evolutionary job allocation {components}, including the individual representation, population initialization, crossover and mutation operators, and fitness evaluation.

\subsubsection{Individual Representation} 
\begin{figure}[hbtp]
    \centering
    \includegraphics[width=0.45\textwidth]{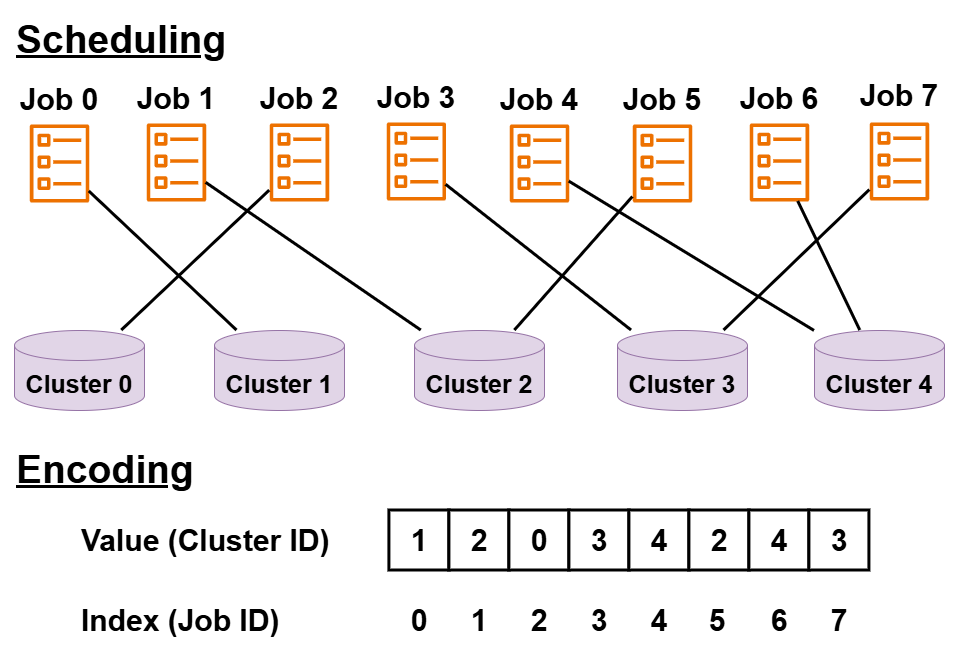} 
    \caption{An example of an individual encoding and its corresponding scheduling.}
    \label{encoding}
\end{figure}

We represent an individual as a one-dimensional vector of length $|\mathcal{J}_s|$, where each index corresponds to a job and the value denotes the assigned cluster ID. Fig.~\ref{encoding} illustrates an example. Since each job index appears exactly once, the representation satisfies the unique allocation constraints in Eqs. \eqref{c6}--\eqref{c7}.

\subsubsection{Population Initialization} We employ randomized initialization; however, purely random assignments may violate constraint \eqref{c10} (e.g., assigning a large job to an incapable cluster). Such infeasible individuals may prevent the construction of a valid schedule and hinder search. Therefore, when initializing the assigned cluster for job $j$, we first compute the feasible cluster set $\mathcal{P}_{f,j}=\{p\in\mathcal{P}|\left\lfloor\frac{G_p\times r_j}{g_j}\right\rfloor\times N_p\geq r_j\}$, and then randomly select a cluster from $\mathcal{P}_{f,j}$.

\subsubsection{Crossover and Mutation Operators} During offspring generation, adopt a single-point crossover operator~\cite{survey-crossover}. For mutation, we use two complementary strategies: (i) single-job mutation, which reassigns one randomly selected job to another feasible cluster, and (ii) swap-based mutation, which exchanges the assigned clusters of two jobs. One strategy is selected uniformly at random each time mutation is applied. The mutation operator must preserve feasibility with respect to \eqref{c10}. For the single-job mutation, the new cluster is sampled from the job’s feasible set. For the swap mutation, given a solution, we randomly select a job $j$ with current cluster $p$ and its feasible cluster set $\mathcal{P}_{f,j}$. Let $\mathcal{J}_a$ denote the set of jobs assigned to clusters in $\mathcal{P}_{f,j}$. We then filter $\mathcal{J}_a$ to obtain jobs that can run on cluster $p$:
$\mathcal{J}_f = \{ w \in \mathcal{J}_a \mid \left\lfloor \frac{G_p \times r_w}{g_w} \right\rfloor \times N_p \ge r_w\}$.
Finally, we randomly select one job from $\mathcal{J}_f$ and swap its cluster assignment with that of job $j$, ensuring the resulting solution remains feasible.


\subsubsection{Fitness Evaluation} Each individual encodes only a job-to-cluster mapping; thus, we invoke the resource allocation procedure to construct a complete schedule before computing the objective value. We use Eq.~\eqref{objective} as the fitness function, so the evolutionary job allocation directly optimizes the scalarized multi-objective formulation.


\subsection{Details for Resource Allocation with Priority Function Learned by GP (RAPFTGP)}
\label{sec:resource_allocation}



When GPU resources are insufficient to satisfy concurrent job demands, the scheduler must resolve contention by prioritizing jobs. Such policies can be expressed as a \emph{priority function}, which maps job features to a scalar score used for ranking. Prior work shows that priority function design critically affects ML scheduling performance~\cite{method-tiresias}. To automatically derive effective and interpretable rules, we use GP to learn a priority function offline.

Algorithm~\ref{resourceAllocation} describes the resource allocation procedure. The inputs include an individual $c$ (job-to-cluster mapping), the schedulable job set $\mathcal{J}_s$, the non-preemptive job set $\mathcal{J}_n$, the cluster set $\mathcal{P}$, and the priority function $f(\cdot)$.

\begin{algorithm}[hbtp]
  \SetKwData{Left}{left}\SetKwData{This}{this}\SetKwData{Up}{up}
  \SetKwFunction{Union}{Union}\SetKwFunction{FindCompress}{FindCompress}
  \SetKwInOut{Input}{Input}\SetKwInOut{Output}{Output}

  \Input{Individual $i$, The set of jobs to be scheduled
 $\mathcal{J}_s$, non-preemptive job set $\mathcal{J}_n$, cluster set $\mathcal{P}$, Priority function $f(\cdot)$.}
  \Output{Solution $s$.}
  $\mathcal{X}\leftarrow\emptyset$, $\mathcal{Y}\leftarrow\emptyset$\;
  \For{$p\in\mathcal{P}$}{
    $\mathcal{J}_{p,n}\leftarrow\{j\in\mathcal{J}_n|c_j=p\}$\;
    $\mathcal{J}_{p,s}\leftarrow\mathcal\{j\in \mathcal{J}_s|i[j.id]=p\}$\;
    $y_p\leftarrow$ An empty list\;
    \tcp{Handling Non-preemptive Jobs}
    \For{$j\in\mathcal{J}_{p,n}$}{
        \For{$n\in  \mathcal{N}_p$}{
            $x_{j,n}\leftarrow w_{j,n}$\;
            $\mathcal{X}\leftarrow\mathcal{X}\cup\{x_{j,n}\}$\;
        }
        Append $j$ to $y_p$;
    }
    \tcp{Handling Schedulable Jobs}
    \While{$|\mathcal{J}_{p,s}|>0$}{
        $j\leftarrow$ The job with the minimum $f(j)$ in $\mathcal{J}_{p,s}$\;
        {$t_{j,s}\leftarrow $ Find the earliest feasible start time of $j$ based on already scheduled jobs\;}
        \tcp{Best Fit Allocation}
        \For{{each worker of job $j$}}{
            $n\leftarrow \arg\min\limits_{n\in\mathcal{N}_p}\{{\sigma}(n,t_{j,s})|{\sigma}(n,t_{j,s})\geq\frac{g_j}{r_j}\}$\;
            $x_{j,n}\leftarrow x_{j,n}+1$\;
            $\mathcal{X}\leftarrow \mathcal{X}\cup\{x_{j,n}\}$
        }
        Append $j$ to $y_p$\;
        $\mathcal{J}_{p,s}\leftarrow \mathcal{J}_{p,s}\setminus\{j\}$\;
    }
    $\mathcal{Y}\leftarrow\mathcal{Y}\cup\{y_p\}$\;
  }
  $s\leftarrow (\mathcal{X}, \mathcal{Y})$\;
  \caption{Resource Allocation}
  \label{resourceAllocation}
\end{algorithm}


The algorithm initializes $\mathcal{X}$ to record worker placements and $\mathcal{Y}$ to record intra-cluster job orderings. It then processes clusters sequentially. For each cluster, it first fixes non-preemptive jobs by preserving their placements and placing them at the front of the ordering. {Note that the relative order among non-preemptive jobs does not affect the scheduling outcome, since they are already running at time $t$ and their resource allocations are fixed} (lines 5--12). For schedulable jobs, it ranks jobs using $f(\cdot)$ and allocates resources in that order (line 14). For each job, it computes the earliest feasible start time and then applies Best Fit to place each worker (lines 15--20). 
The earliest start time of job $j$ is computed according to Eq.~\eqref{start}.
\begin{equation}
\label{start}
    \begin{aligned}
        t_{j,s} = \min \left\{ t_{j,s} \in \mathcal{T} \,\middle|\,
        \sum_{n \in \mathcal{N}_p}
        \left\lfloor \frac{\sigma(n,t_{j,s})}{g_j / r_j} \right\rfloor
        \ge r_j \right\}, \\
        \mathcal{T} = \{t\} \cup \{ F_w \mid w \in {\mathcal{Y}_{p,<j}} \}.
    \end{aligned}
\end{equation}
{where $\mathcal{Y}_{p,<j}$ denotes the set of jobs in cluster $p$ that have already been scheduled before job $j$ according to the priority order (i.e., the jobs that have already been added into $y_p$ when job $j$ is being scheduled), and the candidate time set $\mathcal{T}$ consists of the current time $t$ and the completion times of these jobs, since resource availability only changes at these time instants. 
For each job $w$, its completion time is given by $F_w = t_{w,s} + d_w^{(t)}$. 
Moreover, $\sigma(n, t_{j,s})$ denotes the remaining capacity of node $n \in \mathcal{N}_p$ at time $t_{j,s}$ under the current scheduling decision.}

The remaining capacity of node $n$ at time $t_{j,s}$ can be calculated as
\begin{equation}
\label{cap}
    \sigma(n,t_{j,s})
    = G_p 
    - \sum_{w \in \mathcal{Y}_{p.<j}} 
    \left(
        x_{w,n} \cdot \boldsymbol{1}_{F_w > t_{j,s}} \cdot \frac{g_w}{r_w}
    \right).
\end{equation}
After completing the resource allocation for all clusters, $(\mathcal{X}, \mathcal{Y})$ is output as the complete solution (line~26).

Algorithm~\ref{RAPFTGP} presents the GP procedure for learning the priority function. The inputs include a {training problem instance set $\mathcal{D}$}, a {validation problem instance set $\mathcal{V}$}, population size $n$, mutation rate $r$, and training subset size $\alpha$. {Both $\mathcal{D}$ and $\mathcal{V}$ consist of multiple problem instances, where each instance composed of a set of machine learning jobs and a set of clusters.} In each generation {of GP}, a subset $\mathcal{D}_{sub}$ is sampled for fitness evaluation~\cite{method-subsetselection}; offspring are generated via subtree crossover and mutation; and the best individuals are selected for the next generation. After termination, individuals are evaluated on $\mathcal{V}$ and the best one is converted into a human-readable expression, yielding the priority function $f(\cdot)$ used in Algorithm~\ref{resourceAllocation}. The detailed implementation for learning the priority function with GP is presented below.

\begin{algorithm}[htbp]
  \SetKwData{Left}{left}\SetKwData{This}{this}\SetKwData{Up}{up}
  \SetKwFunction{Union}{Union}\SetKwFunction{FindCompress}{FindCompress}
  \SetKwInOut{Input}{Input}\SetKwInOut{Output}{Output}

  \Input{{Training problem instance set $\mathcal{D}$}, {Validation problem instance set $\mathcal{V}$}, Population size $n$, Mutation rate $r$, The size of the training subset $\alpha$.}
  \Output{Score Function $f(\cdot)$}
  ${\mathcal{I}} \leftarrow$ Initialize population\;
  \While{termination conditions are not met}{
    $\mathcal{O}\leftarrow\emptyset$\;
    $\mathcal{D}_{sub}\leftarrow$ Randomly select $\alpha$ instances from $\mathcal{D}$\;
    \For{$|\mathcal{O}|<n$}{
        $p_1, p_2 \leftarrow$ Select from $\mathcal{I}$ with $k$-tournament selection approach\;
        $c_1, c_2\leftarrow$ Offspring generated by crossover of $p_1$ and $p_2$\;
        \If{$rand(0,1)<r$}{
            Apply the mutation operation to individual $c_1$ and $c_2$.
        }
        $\mathcal{O}\leftarrow\mathcal{O}\cup\{c_1,c_2\}$\;
    }
    \For{${i}\in \mathcal{I}\cup\mathcal{O}$}{
        Evaluate the fitness of $i$ with $\mathcal{D}_{sub}$\;
    }
    $\mathcal{I}\leftarrow$ Select $n$ individuals with the best fitness\;
  }
  \For{${i}\in \mathcal{I}$}{
    Evaluate the fitness of $i$ with $\mathcal{V}$\;
  }
  $r_{best}\leftarrow$ Select the best individual\;
  $f(\cdot) \leftarrow$ Convert individual $r_{best}$ into priority function\;
  \caption{Priority Function Learning with GP}
  \label{RAPFTGP}
\end{algorithm}

\subsubsection{Individual Representation}
In the algorithm, individuals are represented using a tree structure (binary tree), where the internal nodes are selected from the function set, and the leaf nodes are selected from the terminal set. This representation can be directly converted into prefix, infix, or postfix expressions, yielding an interpretable priority function~\cite{method-accgphhafolraicbc}.

\subsubsection{Function Set and Terminal Set}
Table~\ref{term&Func} summarizes the terminal and function sets. We include remaining duration $d_j^{(t)}$ and total GPU demand $g_j$ to capture temporal and resource dimensions~\cite{method-tiresias}, and include queuing time $q_j^{(t)}$ and user priority $u_j$ to align with the optimization objectives. The function set is $\{+, -, \times, \%\}$, where $\%$  denotes protected division~\cite{book-GP}.  If the denominator is zero, the numerator is returned directly; otherwise, the division operation is performed.


\begin{table}[htbp]
\caption{Function and Terminal sets of GP}
\centering
\begin{tabular}{ll} 
\toprule 
Terminal Set & Description\\
\midrule
$d_j^{(t)}$ & The remaining duration of job $j$.\\
$q_j^{(t)}$ & The length of time that job j has been queued.\\
$u_j$ & The priority defined by the user for job $j$.\\
$g_j$ & The total number of GPUs required by job $j$.\\
\addlinespace[3pt]
\toprule
Function Set & $+, -, \times, \%$ (protected division)\\
\bottomrule
\end{tabular}
\label{term&Func}
\end{table}

\subsubsection{Fitness Evaluation}
The fitness of a GP individual should reflect the scheduling quality induced by its priority function. Thus, for each problem instance (jobs and cluster resources), we construct a schedule using Algorithm~\ref{resourceAllocation} with the candidate priority function and compute the resulting objective value. To reduce learning cost, we use the Minimum Completion Time (MCT) heuristic~\cite{method-multicloud} for job-to-cluster assignment during priority function learning, while keeping the intra-cluster resource allocation identical. This choice avoids the large computational overhead of running evolutionary job allocation for each GP fitness evaluation, while still exposing GP to representative scheduling dynamics.

\begin{figure}[hbtp]
    \centering
    \includegraphics[width=0.45\textwidth]{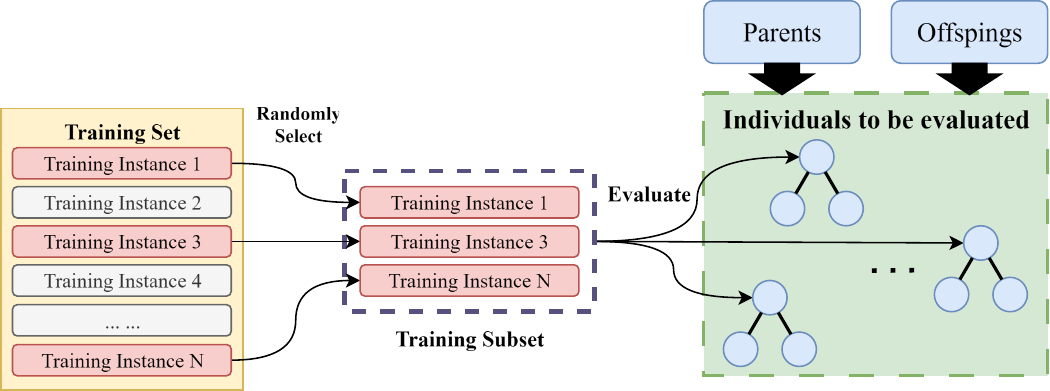} 
    \caption{A training subset is constructed within each generation for fitness evaluation.}
    \label{subset}
\end{figure}

In the fitness evaluation process of GP, a common method is to evaluate each individual on all training problem instances~\cite{method-accgphhafolraicbc}. However, when the training problem instance set is large, this becomes expensive. Inspired by~\cite{method-subsetselection}, we construct a compact training subset in each generation for fitness evaluation (Fig.~\ref{subset}).

The fitness of an individual is computed as the sum of its scores over instances in $\mathcal{T}_{sub}$:
\begin{equation}
    fitness(c) = \sum_{i\in\mathcal{T}_{sub}}score(c, i)
    \label{fitness}
\end{equation}
where $score(c, i)$ represents the score of individual $c$ on the training instance $i$. 
To reduce variance across instances, we normalize objective values~\cite{method-accgphhafolraicbc},
\begin{equation}
    score(c, i) = \frac{O_{c, i}}{O_{base, i}}
    \label{score}
\end{equation}
where $O_{c,i}$ is the objective value on problem instance $i$ using the priority function of individual $c$, and $O_{base,i}$ is the objective value using a baseline priority function. We use SRSF~\cite{method-tiresias} as the baseline, {whose priority function is denoted by $f(j)=g_j\cdot d_j^{(t)}$} (smaller values indicate higher priority), based on its stable performance across workload pressures observed in our experiments.


\section{Experimental Studies}
\label{sec:exp}
In this section, we empirically evaluate EVO-GP in both multi-cluster and single-cluster GPU environments.
{Although the primary focus of this paper is multi-cluster scheduling, we additionally include single-cluster experiments to demonstrate that EVO-GP can be adapted beyond the specific objective set and environment considered in this work.}
We first describe the experimental settings, including workloads, cluster configurations, performance metrics, and baselines. We then present results in three aspects:
(i) overall performance in multi-cluster scenarios,
(ii) performance and adaptability in single-cluster scenarios, and
(iii) robustness under varying load pressures.
Finally, we conduct ablation studies to quantify the contribution of each component {in our proposed framework} and report the online running time overhead of EVO-GP. {Unless otherwise stated, all experimental results reported in this section are obtained by averaging over 15 independent runs with different random seeds.}
\subsection{Experimental Setting}

\subsubsection{Workload Types} 
We conduct experiments on two types of workloads: (i) real-world job traces collected from SenseTime’s production cluster~\cite{trace-capodlwilsgd}, and (ii) a workload simulator constructed through trajectory analysis of Huawei’s production cluster. Both workloads correspond to online settings, where jobs arrive sequentially and the scheduler is triggered upon each job arrival.
\begin{itemize}
    \item \textbf{SenseTime Traces}~\cite{trace-capodlwilsgd}: To evaluate generality {of our framework} on real traces and {conventional} objectives, we use data from SenseTime’s Venus cluster. Specifically, we use a two-week snapshot containing 9,281 jobs on a cluster with 1,064 GPUs as the test set, and the preceding two-week snapshot containing 9,820 jobs as the training set.     
    \item \textbf{Workload simulator}: We employ the simulator provided by Huawei to evaluate EVO-GP under controllable multi-cluster configurations. The simulator models 21 job types, summarized in Table~\ref{workload}. The first two columns specify the required number of workers and GPUs for each type. We consider synchronous data-parallel training, where GPUs per worker are fixed as $\frac{\text{GPU Num}}{\text{Worker Num}}$. The third column reports the mean training time (hours); for each type, training times follow an exponential distribution with the given mean. The simulator uses a 1-second time unit (fractional seconds are discarded). {The last column ``Weight" specifies the relative sampling probability of each job type in the workload simulator, reflecting its occurrence frequency in the production traces.} For each cluster, the simulator first filters out infeasible job types and then samples types according to their weights. Job priorities are sampled uniformly from $\{1,2,3\}$. Job arrivals follow a Poisson process with rate $\lambda$. In our simulations, the total number of arrivals per cluster is set to $10\times N_p$~\cite{simulator1}, where $N_p$ denotes the number of nodes in cluster $p$. For each cluster $p$, $\lambda$ is sampled uniformly from $[\lambda_{\min}, \lambda_{\max}]$, where $ \lambda_{\max} = \frac{N_p\times G_p}{\mathbb{E}[d_j\times g_j]}$ and $\lambda_{\min} = \frac{\lambda_{\max}}{4}$~\cite{simulator1,simulator2}.
\end{itemize}

\begin{table}[hbtp]
\centering
\caption{Types of Machine Learning Jobs and Their Characteristics}
\begin{tabular}{lccc}
\toprule
\textbf{Worker Num} & \textbf{GPU Num} & \textbf{Avg. Training Hours} & \textbf{Weight} \\
\midrule
1  & 1    & 1.499136  & 2.44 $\times$ 10$^{-1}$ \\
1  & 2    & 0.906435  & 1.23 $\times$ 10$^{-1}$ \\
1  & 4    & 1.050502  & 8.20 $\times$ 10$^{-2}$ \\
1  & 8    & 3.285060  & 4.29 $\times$ 10$^{-1}$ \\
2  & 2    & 1.701724  & 3.00 $\times$ 10$^{-2}$ \\
2  & 4    & 1.348965  & 1.00 $\times$ 10$^{-3}$ \\
2  & 8    & 0.167778  & 1.80 $\times$ 10$^{-4}$ \\
2  & 16   & 6.448247  & 2.40 $\times$ 10$^{-2}$ \\
3  & 24   & 0.277222  & 3.70 $\times$ 10$^{-4}$ \\
4 & 4    & 0.039333  & 2.64 $\times$ 10$^{-4}$ \\
4 & 32   & 4.548746  & 2.90 $\times$ 10$^{-2}$ \\
6 & 48   & 5.659565  & 3.30 $\times$ 10$^{-3}$ \\
8 & 8    & 0.751250  & 4.00 $\times$ 10$^{-4}$ \\
8 & 64   & 11.871677 & 1.93 $\times$ 10$^{-2}$ \\
10 & 80   & 6.448889  & 8.50 $\times$ 10$^{-4}$ \\
12 & 96   & 0.449074  & 1.70 $\times$ 10$^{-4}$ \\
16 & 16   & 0.135417  & 1.00 $\times$ 10$^{-3}$ \\
16 & 64   & 0.036111  & 5.30 $\times$ 10$^{-5}$ \\
16 & 128  & 12.898697 & 3.91 $\times$ 10$^{-2}$ \\
24 & 192  & 0.198611  & 5.30 $\times$ 10$^{-5}$ \\
32 & 256  & 12.730382 & 8.60 $\times$ 10$^{-4}$ \\
\bottomrule
\end{tabular}
\label{workload}
\end{table}

\subsubsection{Cluster Resource Model} 
In multi-cluster experiments, each test instance consists of multiple clusters with varying numbers of nodes. The number of clusters is sampled from $[2,26]$, following the virtual-cluster settings reported in SenseTime~\cite{trace-capodlwilsgd}. The number of nodes per cluster is sampled from $[10,100]$, and each node is equipped with 8 GPUs. For the SenseTime Venus trace, we follow~\cite{method-astraea,scaledown1} and scale down the cluster capacity to 800 GPUs to reduce the impact of CPU-intensive workloads.


\subsubsection{Performance Metric} To evaluate the effectiveness of our algorithm, we conduct experiments in two parts. The first part uses the simulator to test multi-cluster scheduling under the objectives defined in Eqs.~\eqref{AvgWQT}--\eqref{utilization}. The second part uses the SenseTime trace to evaluate EVO-GP in a single-cluster setting under commonly used metrics, demonstrating that our approach can be adapted beyond the specific objective set used in the simulator.


For multi-cluster experiments, we report:
\begin{itemize}
    \item \textbf{Average Weighted Queuing Time} (Eq.~\eqref{AvgWQT}): encourages shorter waiting times, especially for high-priority jobs.
    \item \textbf{Maximum Weighted Queuing Time} (Eq.~\eqref{MaxWQT}): captures worst-case delay and mitigates starvation.
    \item \textbf{GPU Time Utilization} (Eq.~\eqref{utilization}): measures the efficiency of GPU-time usage.
\end{itemize}

For the single-cluster experiments, we report:
\begin{itemize}
    \item \textbf{Average Job Completion Time {(AvgJCT)}}: a {conventional} metric for ML cluster scheduling.
    \item \textbf{Maximum Queuing Time  {(MaxJCT)}}: reflects tail latency in waiting time; it often conflicts with optimizing Average Job Completion Time because policies favoring short jobs can delay large jobs.
\end{itemize}



\subsubsection{Baselines and Implementation Details} 

We compare EVO-GP with widely used schedulers, including (i) Spark~\cite{method-spark}, (ii) Liquid~\cite{method-liquid}, (iii) Tiresias~\cite{method-tiresias}, and (iv) Lyra~\cite{method-lyra}. Spark uses Round Robin for placement, while the other methods use Best Fit placement. Spark~\cite{method-spark} and Liquid~\cite{method-liquid} apply FIFO ordering. When job durations are known, Tiresias~\cite{method-tiresias} uses SRSF to rank jobs; Lyra applies Best-Fit Decreasing. Since dedicated multi-cluster allocation baselines are limited, we additionally use the heuristic Minimum Completion Time (MCT) method~\cite{method-multicloud} as a baseline cluster-allocation strategy.


The specific parameter configurations for EVO-GP are set as follows. For the evolutionary job allocation layer, we set the population size to 100, the maximum number of generations to 200, and the mutation rate to 0.1. For GP-based priority function learning, we initialize the population using the ramped half-and-half method~\cite{method-gp}. The training set contains 50 simulator-generated instances, and the per-generation training subset size is set to 3. The remaining GP hyperparameters follow~\cite{method-accgphhafolraicbc}: population size 512, maximum generations 100, mutation rate 0.1, and maximum tree depth 7.

\subsection{Performance of the Algorithm in a Multi-Cluster Scenario}

\begin{figure}[hbtp]
    \centering
    \includegraphics[width=0.5\textwidth]{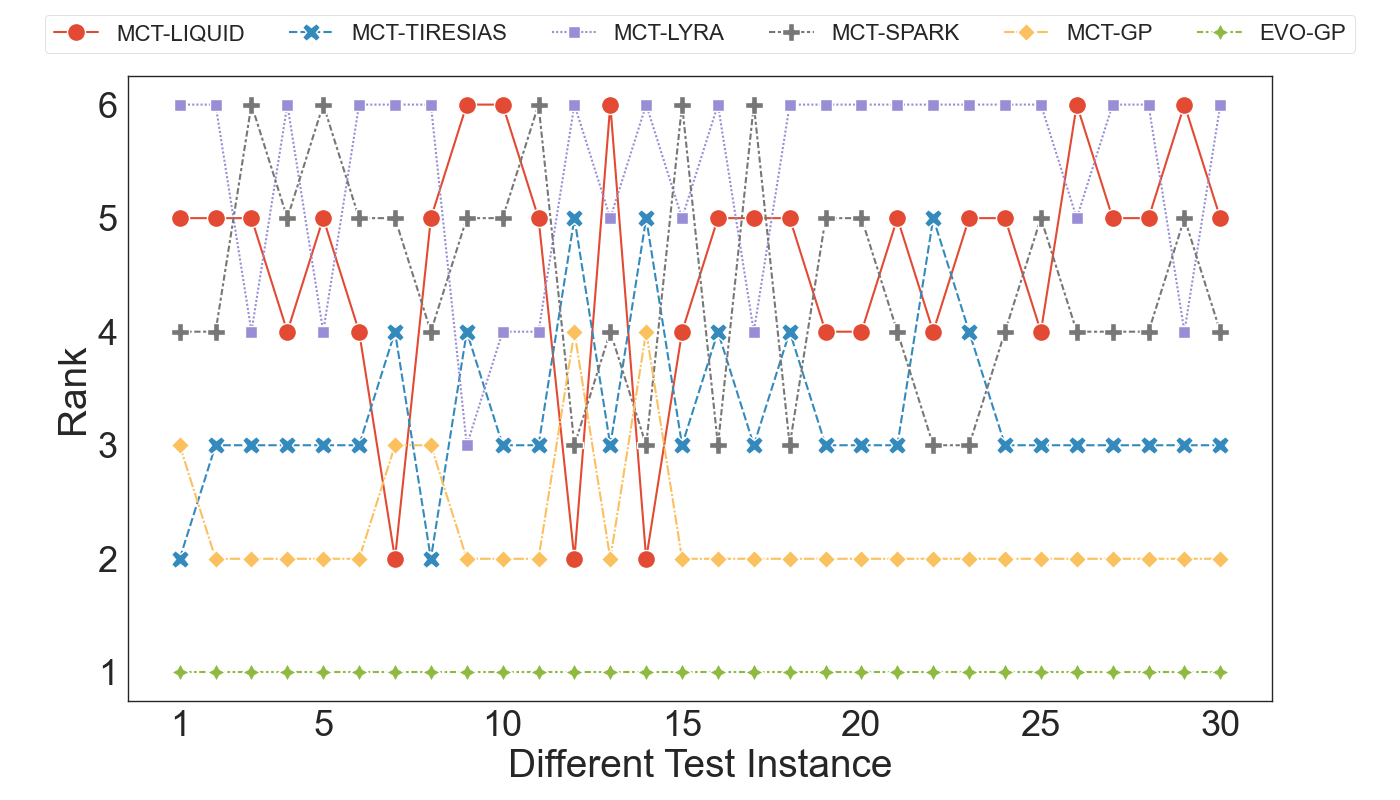} 
    \caption{Ranking of scheduling algorithms under the scalarized multi-objective metric on 30 test instances}
    \label{rank}
\end{figure}

We compare EVO-GP against the baselines in multi-cluster settings. We use Eq.~\eqref{objective} (weighted sum of the three objectives) as the primary evaluation metric, with equal weights ($\omega_1=\omega_2=\omega_3=1$). Fig.~\ref{rank} reports the ranking results on 30 test instances. We denote by MCT-LIQUID, MCT-TIRESIAS, MCT-SPARK, and MCT-LYRA the combinations of MCT job allocation with each single-cluster scheduler. MCT-GP uses MCT for job allocation and the GP-learned priority function for intra-cluster resource allocation. EVO-GP uses evolutionary job allocation with the GP-learned priority function for resource allocation.

We rank all algorithms on 30 instances. As shown in Fig.~\ref{rank}, with MCT allocation, the GP-learned priority function achieves the best performance on 25 instances, second on 3 instances, and third on 2 instances. When both allocation and resource scheduling are optimized by EVO-GP, our method achieves the best performance across all instances.

Experimental results indicate that scheduling algorithms such as Spark~\cite{method-spark} and Liquid~\cite{method-liquid}, which rely on FIFO-based strategies with relative priority ordering, exhibit larger performance variance under {the given} workload. This is primarily because FIFO {considers} only arrival time and ignores other job characteristics, making it sensitive to workload composition and burstiness. In contrast, Tiresias~\cite{method-tiresias} incorporates both resource demand and execution time in its scheduling decisions, leading to more stable performance and overall stronger effectiveness. However, it strongly favors short  jobs, which can starve large jobs and increase the maximum queuing time, thereby degrading the scalarized objective when {the} tail delay is weighted. Lyra~\cite{method-lyra} favors jobs with larger per-worker GPU demand. Its performance is inferior {to ours}, because Lyra~\cite{method-lyra} is designed to exploit job elasticity, whereas we focus on rigid jobs that cannot make progress without receiving their full requested workers ({Constraint} \eqref{c8}).



\begin{figure}[hbtp]
    \centering
	\vspace{-0.35cm} 
	\subfigtopskip=2pt 
	\subfigbottomskip=2pt
	\subfigcapskip=-5pt 
	\subfigure[Priority Function Learned by GP]{
		\label{level.sub.1}
		\includegraphics[width=0.46\linewidth]{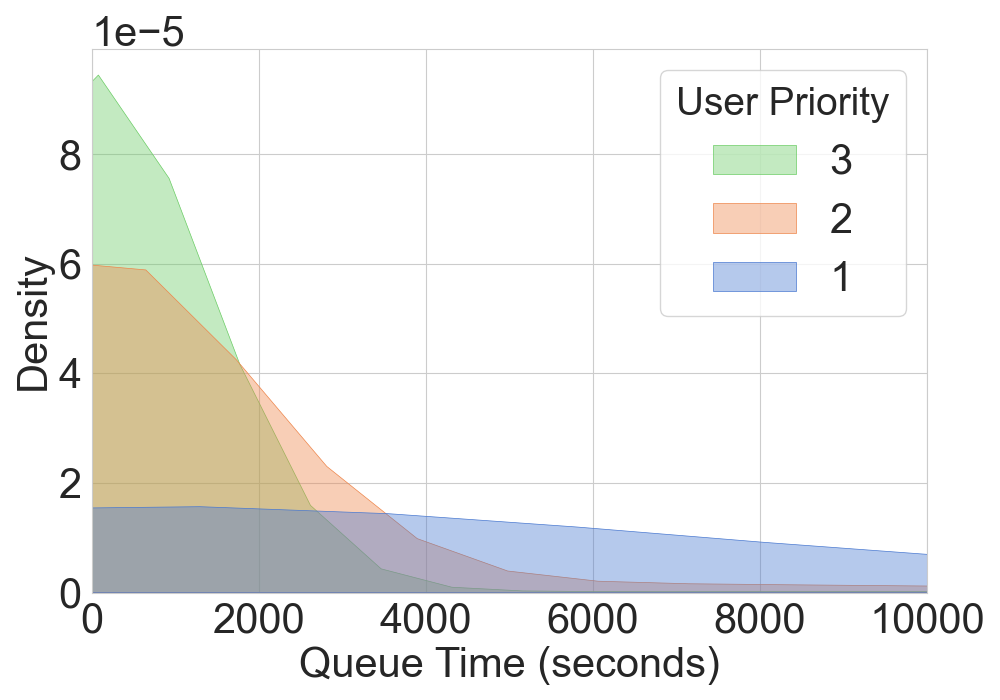}}
	\quad 
	\subfigure[Tiresias Priority Function]{
		\label{level.sub.2}
		\includegraphics[width=0.46\linewidth]{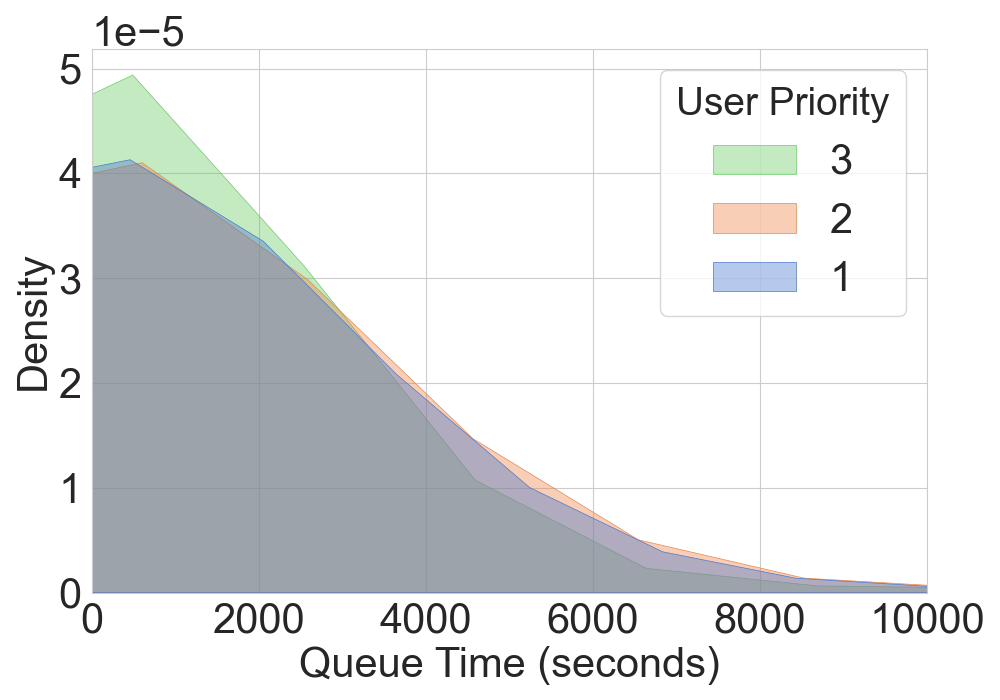}}
	\caption{Job Queuing Time Distributions under GP-Learned and Tiresias Priority Functions.}
	\label{density}
\end{figure}

To further analyze why GP-learned priorities perform well, we conducted simulation experiments on multiple instances and recorded job waiting times across different user-priority levels. Fig.~\ref{density} shows the density distribution of job queuing times on a representative instance with 10 clusters, where the x-axis denotes queuing time and the y-axis denotes probability density. It can be clearly observed that the GP-learned function more aggressively serves higher-priority jobs, yielding a distribution with more probability mass concentrated at smaller queuing times for high-priority classes.


\begin{figure}[hbtp]
	\vspace{-0.35cm} 
	\subfigtopskip=2pt 
	\subfigbottomskip=2pt
	\subfigcapskip=-5pt 
	\subfigure[Avg. Weighted Queuing Time]{
		\includegraphics[width=0.46\linewidth]{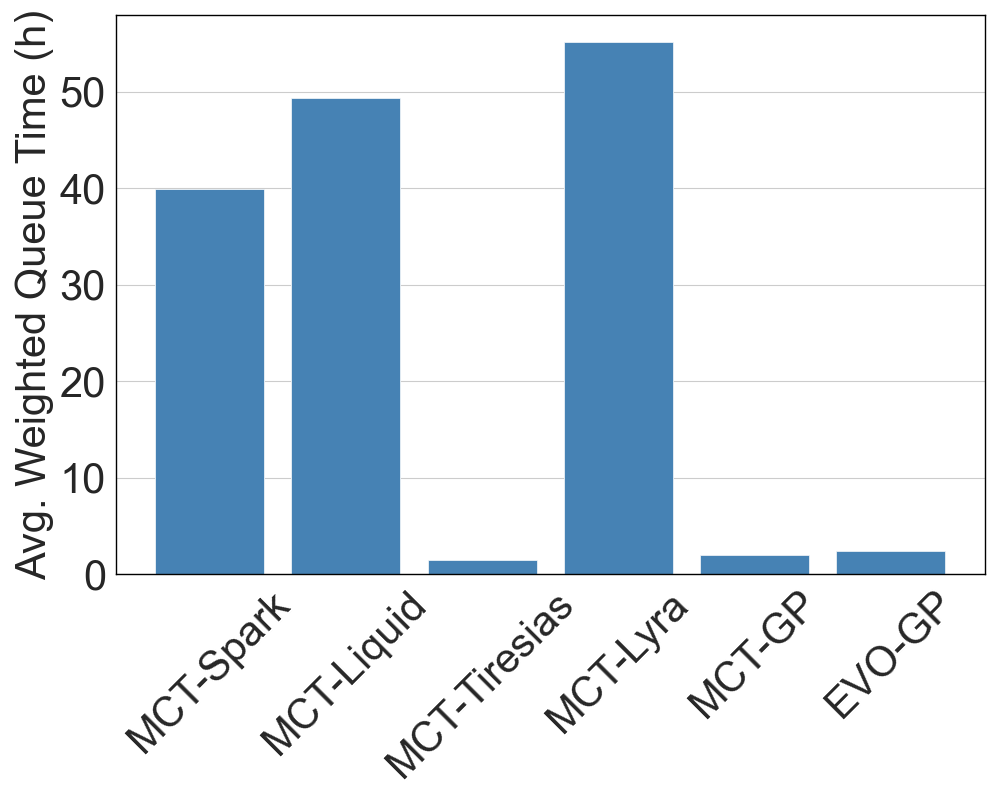}}
	\quad 
	\subfigure[Max Weighted Queuing Time]{
		\includegraphics[width=0.46\linewidth]{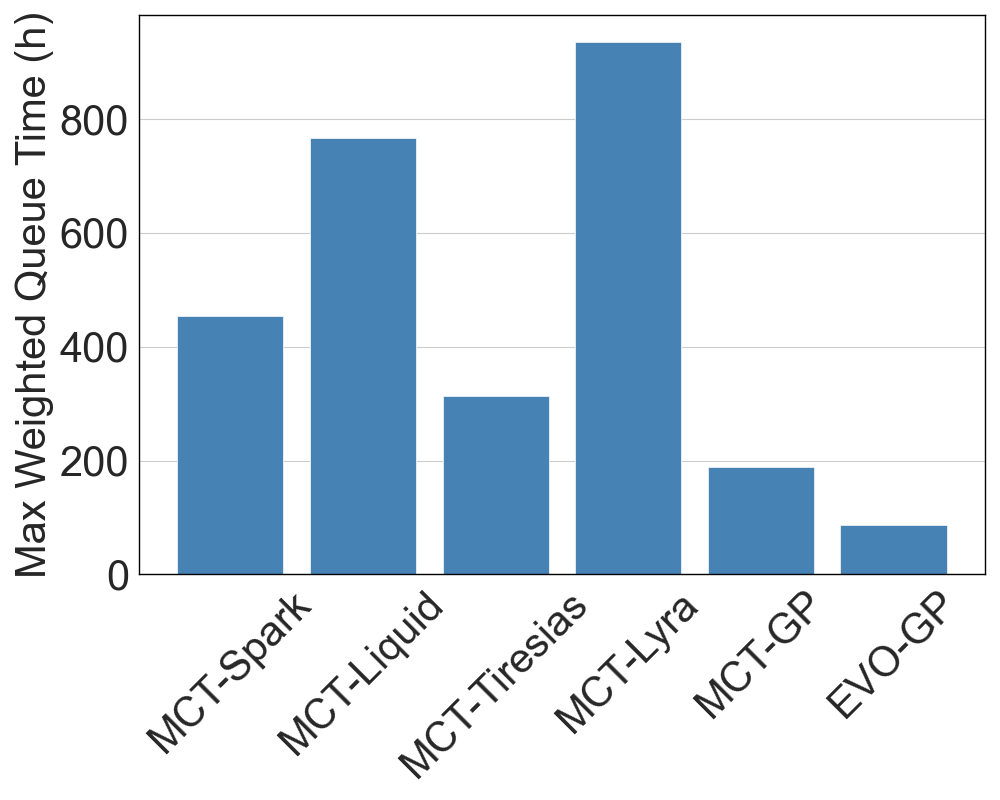}}
    \\
    \subfigure[GPU Time Utilization]{
		\includegraphics[width=0.46\linewidth]{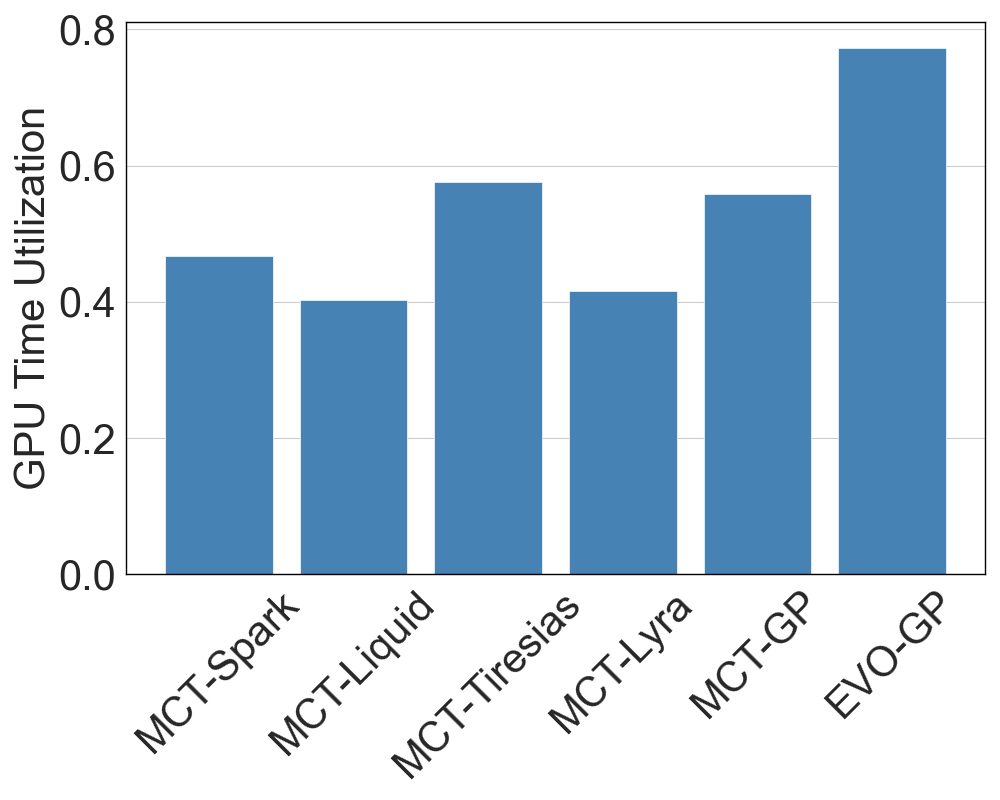}}
    \quad
    \subfigure[Scalarized Objective Value]{
		\includegraphics[width=0.46\linewidth]{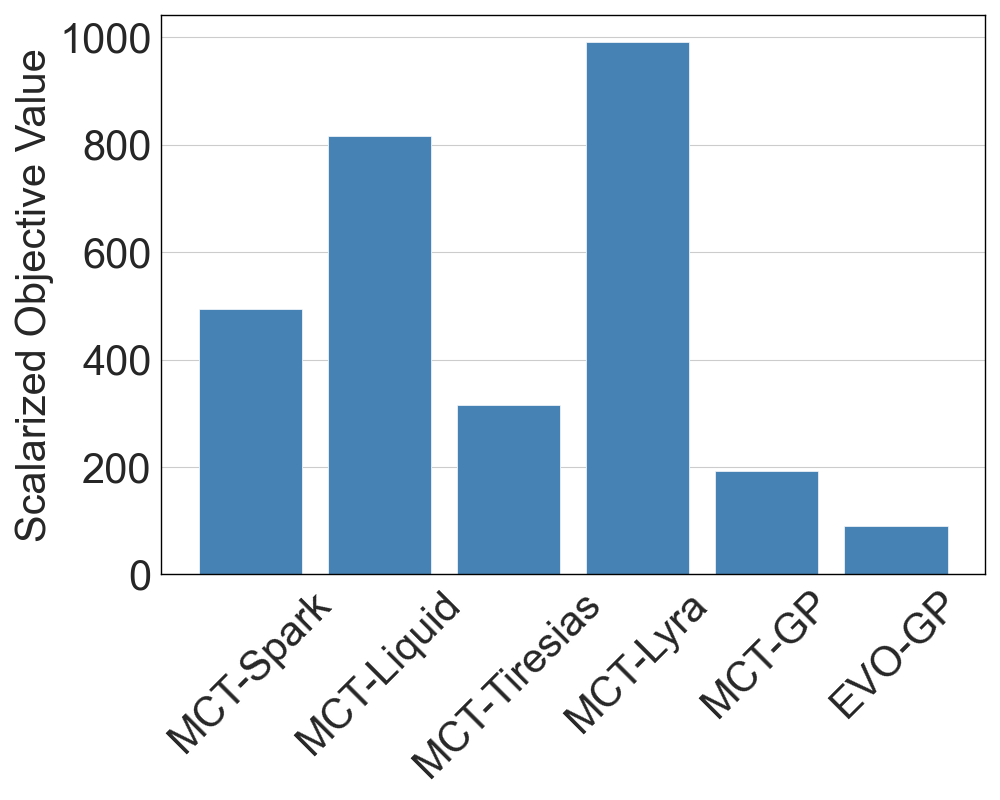}}
	\caption{Different objective values of each algorithm on a representative test instance.}
	\label{objs}
\end{figure}

Fig.~\ref{objs} reports objective breakdowns on representative instances. When all objectives are equally weighted, the maximum weighted queuing time ($C_{q,max}$) often dominates the scalarized objective. The GP-learned priority function therefore emphasizes reducing $C_{q,max}$ while still improving the other objectives when possible. Tiresias~\cite{method-tiresias} achieves strong performance on the average weighted queuing time ($C_{q,avg}$) and GPU time utilization ($C_{u}$) but suffers from a larger $C_{q,max}$. In contrast, the GP-learned function trades off some average weighted queuing time and GPU time utilization to substantially reduce $C_{q,max}$, leading to a better overall scalarized objective.


\subsection{Performance of the Algorithm in a Single-Cluster Scenario}
\label{section-single}
In the previous {subsection}, we demonstrated that the proposed algorithm performs well in scenarios involving multiple clusters with user-specified priorities. In this subsection, we conduct experiments using real-world traces, {according to} the classical objective of minimizing average job completion time and the objective of minimizing maximum completion time. This aims to verify the flexibility of the proposed algorithm across different objectives and scheduling scenarios. The following scalarized objective is used to learn the priority function,


\begin{equation}
    {\rm minimize}\quad \omega \cdot AvgJCT + (1-\omega)\cdot MaxQT
    \label{single-obj}
\end{equation}
where $\omega\in[0,1]$ controls the trade-off. Larger $\omega$ places more emphasis on AvgJCT, while smaller $\omega$ emphasizes MaxQT. 

\begin{figure}[hbtp]
    \centering
    \includegraphics[width=0.4\textwidth]{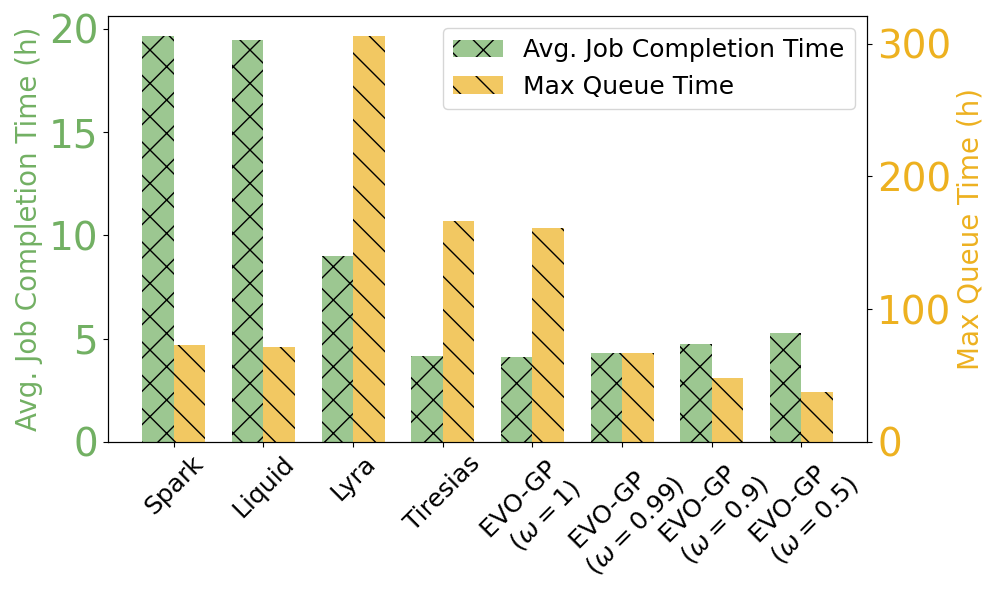} 
    \caption{Results of AvgJCT (Average Job Completion Time) and  MaxQT (Maximum Queuing Time) from various algorithms.}
    \label{all_trend}
\end{figure}

Fig.~\ref{all_trend} shows the results of various algorithms. Tiresias performs well on AvgJCT but yields large MaxQT due to its preference for short jobs. Lyra shows a similar tail-delay issue. Spark and Liquid reduce MaxQT by FIFO ordering but underperform on AvgJCT because they ignore job characteristics. In contrast, the GP-learned priority function adapts to different $\omega$ values and achieves improved trade-offs between AvgJCT and MaxQT.

\begin{figure}[hbtp]
	\subfigure[AvgJCT]{
		\includegraphics[width=0.46\linewidth]{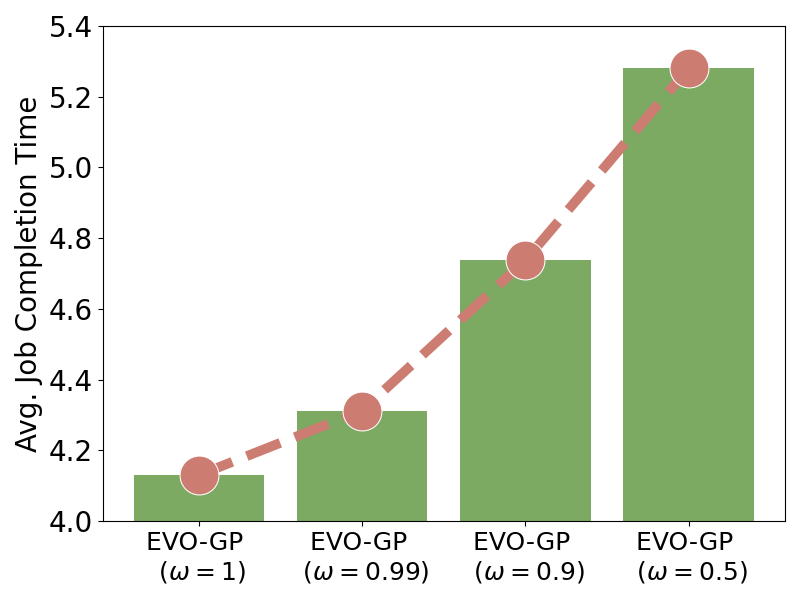}}
	\quad 
	\subfigure[MaxQT]{
		\includegraphics[width=0.46\linewidth]{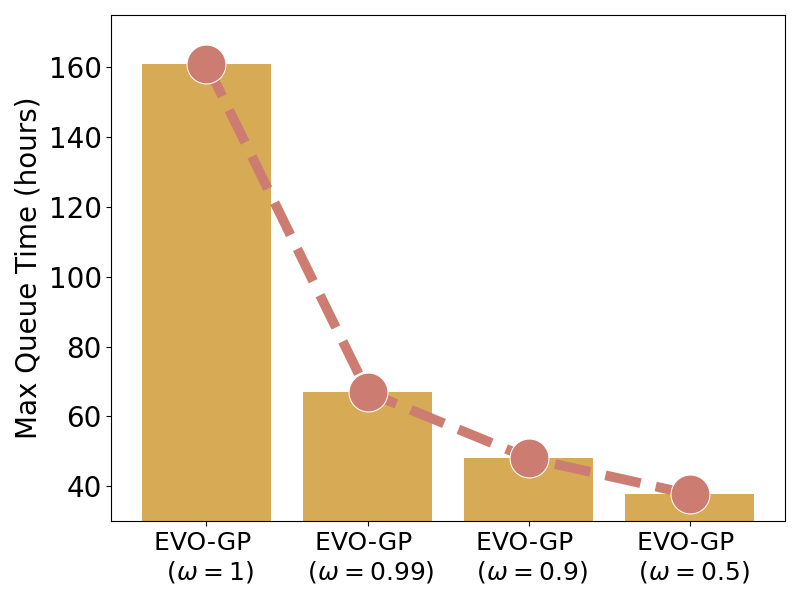}}
	\caption{AvgJCT and MaxQT of the proposed EVO-GP under different values of $\omega$.}
	\label{single_trend}
\end{figure}

Fig.~\ref{single_trend} shows that the optimization behavior of EVO-GP changes under different values of $\omega$. As $\omega$ decreases, the learned priority function places more emphasis on reducing MaxQT, reflecting a controllable and smooth trade-off.


From the experimental results, we can observe that the priority function learned by GP exhibits superior performance compared to expert heuristic rules. This is because GP, as a data-driven approach, can consider problem information more comprehensively and design finer-grained scheduling rules. Additionally, this method can more flexibly adapt to different business scenarios by customizing the objective function, making it more flexible than expert heuristic methods.

\subsection{The Impact of Load Pressure}
\label{sec:pressure}

\begin{figure}[hbtp]
	\subfigure[Simulated workloads]{
        \label{multi_pressure}
		\includegraphics[width=0.46\linewidth]{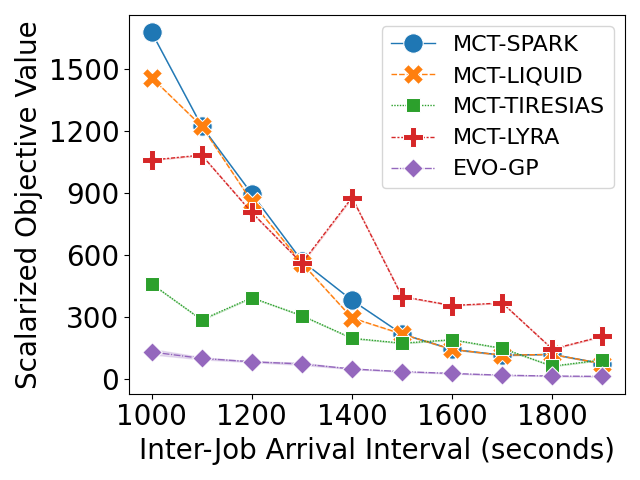}}
	\quad 
	\subfigure[Venus trace]{
        \label{single_pressure}
		\includegraphics[width=0.46\linewidth]{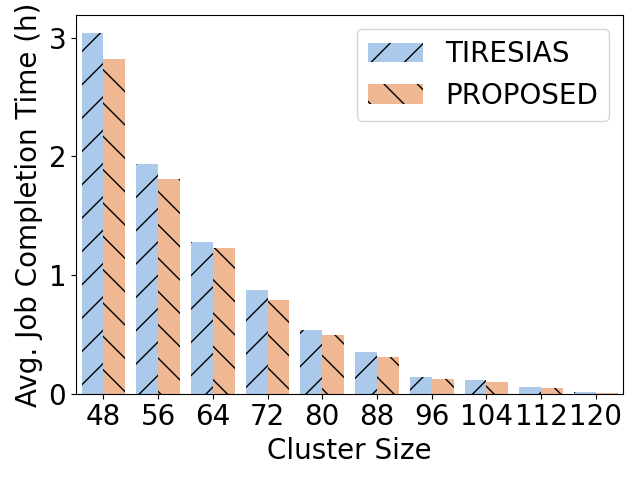}}
	\caption{Impact of load pressure on different algorithms.}
	\label{pressure}
\end{figure}

The level of load pressure also affects the performance of the algorithms. To study robustness under different load pressures, we further conduct experiments under different load pressure in both multi-cluster and single-cluster scenarios. We first conduct a performance comparison of the algorithm in a multi-cluster environment. To better control the load pressure, we assume that the job arrivals follow a uniform distribution \cite{method-DL2}. The load pressure is adjusted by varying the inter-job arrival interval \cite{method-eosfcamlc}. A smaller interval corresponds to higher load pressure, while a larger interval indicates lower load pressure. A total of five clusters are set up, with each cluster containing 40 nodes, and each node equipped with 8 GPUs. Fig. \ref{multi_pressure} presents the experimental results with simulated workloads. It can be observed that EVO-GP consistently outperforms baseline methods under varying load pressures. Moreover, the objective function values achieved by EVO-GP exhibit the minimal fluctuation as the load changes, demonstrating a greater stability compared to other scheduling algorithms.

We then conducted experiments in a single-cluster environment, using the classic objective of AvgJCT for comparison. Following the approach described in \cite{method-astraea}, we controlled the cluster size to adjust the load pressure: a larger cluster corresponds to lower load pressure, while a smaller cluster indicates higher load pressure. During the {priority function learning process}, we construct 10 training instances with different cluster size (shown in Fig. \ref{single_pressure}), each subjected to identical workloads. The training subset size was set to 1, while all other parameters remained consistent with previous settings. To clearly highlight the differences in results, Fig. \ref{single_pressure} only presents a comparison between Tiresias and the proposed method (as both outperform other methods). The results indicate that the priority function learned by GP surpasses that of Tiresias under all load pressure conditions, and this advantage becomes more pronounced as the load increases.

\subsection{{Insight on the Learned Priority Function}}
\begin{figure}[hbtp]
    \centering
    \includegraphics[width=0.48\textwidth]{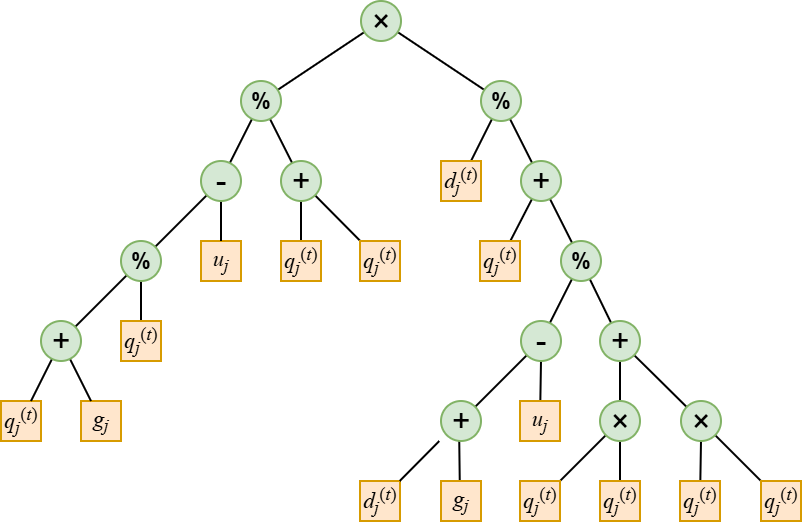} 
    \caption{Example of the learned priority function by GP with workload simulator.}
    \label{insight}
\end{figure}

Fig. \ref{insight} illustrates the tree structure of the priority function learned by GP after simplification. It can be represented as the following mathematical expression:

\begin{equation}
    f(j)=  \frac{\frac{q_j^{(t)}+g_j}{q_j^{(t)}}-u_j}{2q_j^{(t)}}\times\frac{d_j^{(t)}}{q_j^{(t)}+\frac{d_j^{(t)}+g_j-u_j}{(q_j^{(t)})^2}}.
\end{equation}

It is worth noting that a smaller value of $f(j)$ indicates a higher relative priority of job $j$; accordingly, jobs with lower $f(j)$ values are scheduled earlier.
As can be observed, the priority function is expressed as the product of two components. The left component favors jobs with longer queuing times, smaller GPU demands, and higher user-defined priorities, thereby promoting their earlier execution. The right component favors jobs with shorter remaining execution times, longer queuing times, and higher user-defined priorities. These two components preserve the principle of favoring small and short jobs while further incorporating the effects of user-defined priority and queuing time, yielding scheduling decisions that more effectively balance urgency and fairness.

When the queuing time is zero, due to the presence of protected division, Eq. \eqref{insight_w=0} can be further simplified as:
\begin{equation}
    f(j|q_j^{(t)}=0)=\frac{(g_j-u_j)  \times d_j^{(t)}}{d_j^{(t)}+g_j-u_j}
    \label{insight_w=0}
\end{equation}
where the numerator enhances SRSF (i.e., jobs with smaller $g_j\times d_j^{(t)}$ are scheduled earlier) by explicitly incorporating user-defined priority, thereby strengthening the scheduler’s ability to favor high-priority jobs. This term jointly captures a job’s remaining duration, resource demand, and user-defined priority, extending SRSF from a purely efficiency-driven rule to one that simultaneously accounts for urgency and service differentiation.

\subsection{Ablation Study}
\begin{figure}[hbtp]
    \centering
	\vspace{-0.35cm} 
	\subfigtopskip=2pt 
	\subfigbottomskip=2pt
	\subfigcapskip=-5pt 
	\subfigure[Effectiveness of Evolutionary Job Allocation]{
        \label{ablation_evo}
		\includegraphics[width=0.46\linewidth]{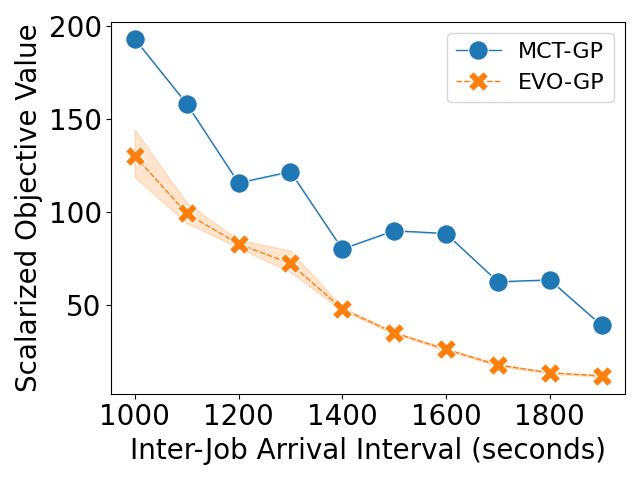}}
	\quad 
	\subfigure[Effectiveness of GP-learned priority function]{
        \label{ablation_gp}
		\includegraphics[width=0.46\linewidth]{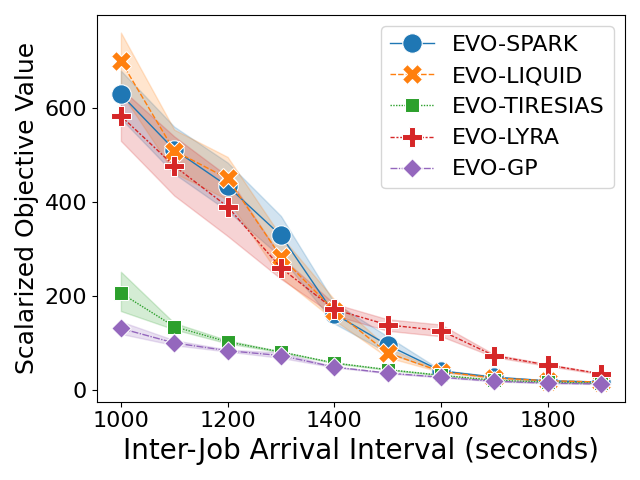}}
	\caption{Ablation experiments: (a) job allocation methods (b) resource allocation methods.}
	\label{ablation}
\end{figure}

To quantify the contribution of each component {of EVO-GP}, we conduct ablation studies in the multi-cluster setting with the same configuration as in Section~\ref{sec:pressure}. Fig. \ref{ablation_evo} compares the performance of job allocation using the evolutionary algorithm and the MCT method under different load pressures. The job allocation using evolutionary algorithm consistently outperforms the MCT method across all load pressures and exhibits smoother trends. Fig. \ref{ablation_gp} compares different intra-cluster resource allocation strategies. It can be observed that when using the GP-learned priority function achieves the best performance with lower variance, especially under heavier load. These results confirm that both the evolutionary allocation layer and the GP-learned priority layer contribute to the overall gains.


\subsection{Algorithm Running Time}
\label{section-time}
\begin{figure}[hbtp]
    \centering
    \includegraphics[width=0.35\textwidth]{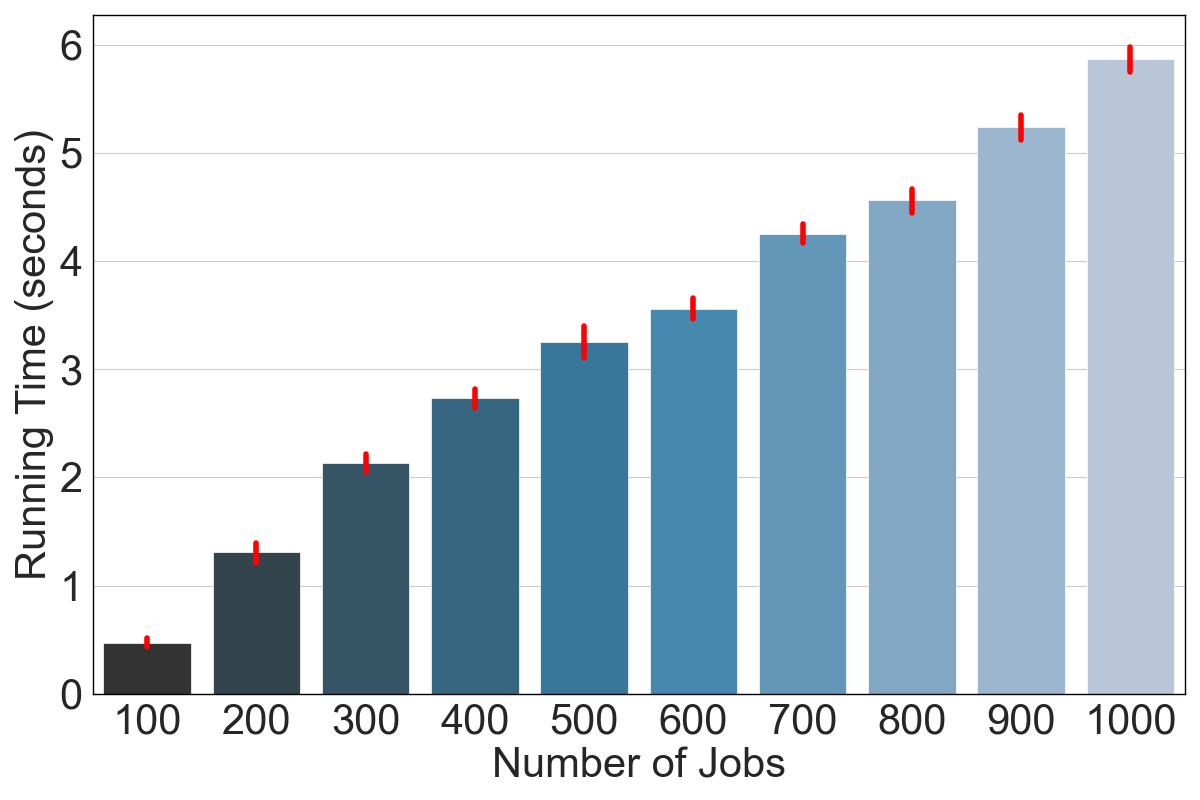} 
    \caption{Running time of EVO-GP to handle different numbers of jobs.}
    \label{runtime_jobs}
\end{figure}

We evaluated the execution time of a single scheduling operation of the EVO-GP online algorithm implemented in JAVA. The experiments were conducted on a desktop computer (Apple M4 Pro, 24 GB RAM) to measure the algorithm’s performance when handling different numbers of jobs. As shown in Fig.~\ref{runtime_jobs}, the execution time increases with the number of jobs but still remains at a relatively low level ($<$ 7 seconds). Considering that ML training jobs typically run for hours or even days, this overhead is acceptable for practical online scheduling.

\section{Conclusion}
\label{sec:conclusion}
This paper proposes EVO-GP, a two-layer evolutionary scheduling framework designed to address the challenges of ML job scheduling in multi-cluster GPU environments. By introducing a two-layer optimization mechanism, the framework decouples global job allocation from local resource scheduling to achieve both efficiency and adaptability. At the job allocation layer, the proposed evolutionary job allocation algorithm dynamically assigns jobs to appropriate clusters to maintain global load balance. At the resource allocation layer, the resource allocation with priority function learned by GP module automatically learns interpretable and effective priority functions that guide GPU allocation decisions under contention. By integrating evolutionary search at the global layer with data-driven, GP-learned heuristics at the local layer, EVO-GP can flexibly adapt to varying workload characteristics and support different multi-objective formulations (e.g., delay, fairness, and utilization). Extensive experiments on both real-world and simulated workloads demonstrate that EVO-GP consistently outperforms state-of-the-art scheduling algorithms in minimizing both average and maximum (weighted) queuing times while maximizing GPU utilization, thereby confirming its effectiveness, scalability, and general applicability in diverse ML job scheduling tasks.

\section*{Acknowledgments}
This work was supported by the National Natural Science Foundation of China (Grant No. 62250710682), the Guangdong Provincial Key Laboratory (Grant No. 2020B121201001), the Program for Guangdong Introducing Innovative and Entrepreneurial Teams (Grant No. 2017ZT07X386), and an internal grant from Lingnan University.

\bibliographystyle{IEEEtran}
\bibliography{references}

\vfill
\end{document}